%% file: main.tex
\documentclass[letterpaper]{article} 
\usepackage[submission]{aaai2026}  
\usepackage{times}  
\usepackage{helvet}  
\usepackage{courier}  
\usepackage[hyphens]{url}  
\usepackage{graphicx} 
\usepackage{natbib}  
\usepackage{caption} 
\usepackage{algorithm}
\usepackage{algorithmic}

\usepackage{newfloat}
\usepackage{listings}
\DeclareCaptionStyle{ruled}{labelfont=normalfont,labelsep=colon,strut=off} 
\floatstyle{ruled}
\newfloat{listing}{tb}{lst}{}
\floatname{listing}{Listing}
\usepackage{amsthm}
\usepackage{svg}
\usepackage{makecell}
\usepackage{booktabs}

\usepackage{dsfont}
\usepackage{amssymb}
\usepackage{amsmath}
\usepackage[misc]{ifsym}
\usepackage{listings}

\title{ExtraGS: Geometric-Aware Trajectory Extrapolation with Uncertainty-Guided Generative Priors}

\author{
~~~~~~Kaiyuan Tan\textsuperscript{\rm \ 1, 2}
~~~~~~Yingying Shen\textsuperscript{\rm \ 2}
~~~~~~Haohui Zhu\textsuperscript{\rm \ 2}
~~~~~~Zhiwei Zhan\textsuperscript{\rm \ 2}
~~~~~~Shan Zhao\textsuperscript{\rm 2} \\
~~~~~~Mingfei Tu\textsuperscript{\rm 2}
~~~~~~Hongcheng Luo\textsuperscript{\rm 2}
~~~~~~Haiyang Sun\textsuperscript{\rm 2 $\dagger$}
~~~~~~Bing Wang\textsuperscript{\rm 2}~\textsuperscript{\Letter} \\
~~~~~~Guang Chen\textsuperscript{\rm 2}
~~~~~~Hangjun Ye\textsuperscript{\rm 2}
\\
~ ~ \textsuperscript{\rm 1}UIUC
~ ~ \textsuperscript{\rm 2}Xiaomi EV
\\
\small{Project Page: \url{https://wm-research.github.io/extrags/}}
}

\usepackage{bibentry}

\begin{document}
\maketitle


\input{sec/0_abstract}

\let\thefootnote\relax
\footnotetext{
\small
\textsuperscript{$\dagger$} Project lead,
\textsuperscript{\Letter} Corresponding author.}

\input{sec/1_intro}

\input{sec/2_related}

\input{sec/2.5_preliminary}
\input{sec/3_method}

\input{sec/4_experiments}

\input{sec/5_conclusion}
\clearpage
{
    \small
    \bibliography{aaai2026}
}

\end{document}

%% file: sec/0_abstract.tex
\begin{abstract}

Synthesizing extrapolated views from recorded driving logs is critical for simulating driving scenes for autonomous driving vehicles, yet it remains a challenging task. Recent methods leverage generative priors as pseudo ground truth, but often lead to poor geometric consistency and over-smoothed renderings. To address these limitations, we propose ExtraGS, a holistic framework for trajectory extrapolation that integrates both geometric and generative priors. At the core of ExtraGS is a novel Road Surface Gaussian(RSG) representation based on a hybrid Gaussian–Signed Distance Function (SDF) design, and Far Field Gaussians (FFG) that use learnable scaling factors to efficiently handle distant objects. Furthermore, we develop a self-supervised uncertainty estimation framework based on spherical harmonics that enables selective integration of generative priors only where extrapolation artifacts occur. Extensive experiments on multiple datasets, diverse multi-camera setups, and various generative priors demonstrate that ExtraGS significantly enhances the realism and geometric consistency of extrapolated views, while preserving high fidelity along the original trajectory.

\end{abstract}

%% file: sec/1_intro.tex
\section{Introduction} 

With the rapid advancement of end-to-end (E2E) autonomous driving methods, there is a growing demand for scalable, controllable, and domain-gap-free closed-loop simulation environments.

Reconstruction-based approaches~\cite{street_gs, omnire, emernerf, pvg, driving_gaussians}, built on recent development in volumetric rendering techniques~\cite{nerf, 3dgs}, have successfully produced high-fidelity driving scene simulation environments with minimal domain gaps. However, these methods primarily target interpolated novel views and face significant challenges when rendering along extrapolated trajectories (e.g., lane changing). 

More recently, several methods~\cite{sgd, drivedreamer4d, recondreamer, streetcrafter, drivex, freesim} have proposed augmenting driving logs with pseudo ground truth for extrapolated views, generated using pre-trained generative priors guided by various control signals (e.g., colored LiDAR projections, noisy renderings, reference images, and object bounding boxes). While these approaches show promising results, the generated pseudo ground truths remain inherently noisy and often inconsistent with the original trajectory data, leading to blurred or distorted renderings. Furthermore, these generative methods require prior knowledge of the extrapolated trajectory, making it computationally infeasible to generate outputs for all possible future paths in practice. 

This motivates two key questions:

\paragraph{How can we improve extrapolation performance through better scene representation design? } Unlike generative methods, a data-independent approach is more general and avoids expensive diffusion model training. However, this approach remains underexplored. Among scene elements, we identify two key elements that undergo severe distortion at extrapolation. Namely, road surfaces and far objects. 

Road surfaces present unique challenges due to parallel viewing angles and sparse textures, making them difficult to reconstruct with traditional volumetric rendering. They also undergo significant view direction changes during extrapolation. To address these challenges, we propose Road Surface Gaussians (RSG): a hybrid model using a dimension-reduced SDF for geometry and flat 3D Gaussians for appearance.

On the other hand, distant objects span vastly different depth ranges and lack LiDAR coverage, making direct optimization inefficient and prone to overfitting. To tackle this issue, we introduce Far Field Gaussians (FFG) that jointly modulate position and scale through learnable scaling factors, enabling rapid convergence for distant objects. 

\begin{figure}[htb]
    \centering
    \includegraphics[width=0.9\linewidth]{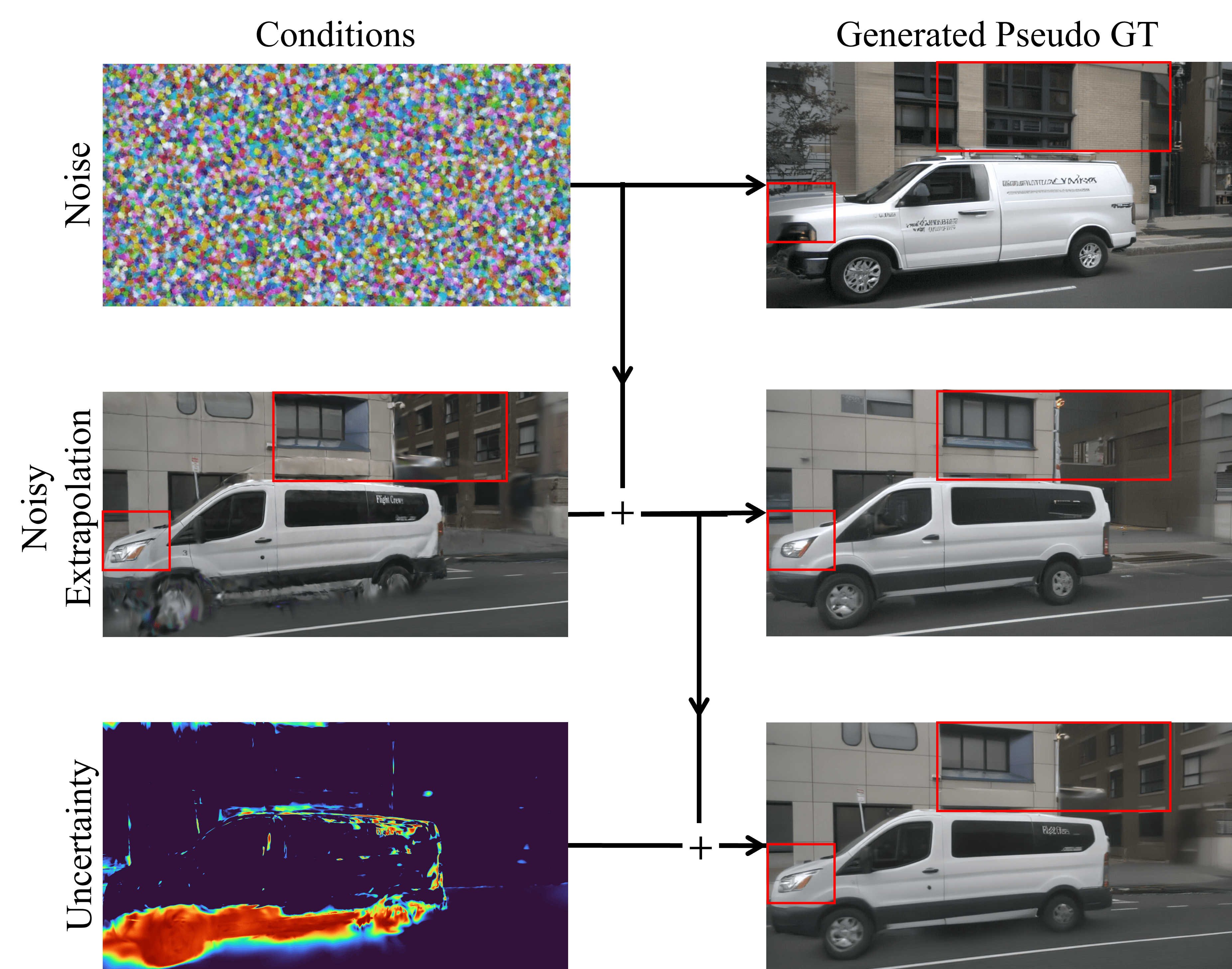}
    \caption{\textbf{Demonstration of Uncertainty-Guided Generation}. \textit{Top:} Pure noise generation produces semantically inconsistent results. \textit{Middle:} Conditioning improves alignment but loses fine details due to uniform pixel treatment. \textit{Bottom:} Our uncertainty-guided approach distinguishes extrapolation-induced noise from scene-inherent noise, enabling targeted generative correction.}
    \label{fig:demo_uncert}
\end{figure}

\paragraph{How can we better leverage generative priors? } While generative models can enhance extrapolation quality, they cannot distinguish between noise induced by viewpoint extrapolation versus scene-inherent characteristics(e.g. intricate patterns). This fundamental limitation leads to either over-smoothing of meaningful scene components(Figure~\ref{fig:demo_uncert}) or preservation of unwanted artifacts(Figure~\ref{fig:ablation_qual}) in extrapolated views. The core challenge lies in identifying which pixels are ``uncertain''--meaning they have a high probability of containing extrapolation-induced noise that generative models should address.
To solve this, we propose a self-supervised uncertainty estimation framework. Our approach seamlessly integrates into the 3DGS framework by storing an additional set of SH coefficients alongside existing view-dependent color representations.
Our contributions can be summarized as follows: 
\begin{enumerate}
    \item We introduce two novel geometric representations for robust extrapolation. Road Surface Gaussians(RSG) incorporates road surface priors, while Far Field Gaussians(FFG) enables efficient optimization of distant objects. 
    \item We propose a self-supervised uncertainty estimation framework based on spherical harmonics that explicitly distinguishes between extrapolation-induced noise and scene-inherent characteristics, allowing precise and selective integration of generative priors.
    \item Through comprehensive experiments across multiple datasets, multi-camera configurations, and generative prior selections, we demonstrate that ExtraGS significantly enhances the realism and structural consistency of extrapolated views. 
\end{enumerate}

%% file: sec/2_related.tex
\section{Related Work}
\label{sec:related}

\paragraph{Novel View Synthesis for Driving Scenes}
Building on volumetric rendering advances~\cite{nerf, 3dgs}, street view synthesis has gained wide-spread attention in recent years. Approaches such as SUDS~\cite{suds}, EmerNeRF~\cite{emernerf}, and PVG~\cite{pvg} handle dynamic objects implicitly through unified representations, where static-dynamic separation emerges during training. While elegant, they treat dynamic objects as undifferentiated wholes, limiting control for simulation applications.
Explicit decomposition methods model dynamic actors separately using object-level poses, enabling manipulation and editing. Early works~\cite{s-nerf, mars, unisim, neurad} based on Neural Scene Graphs train individual NeRFs for each object and background. Recent methods~\cite{driving_gaussians, street_gs, hugs, sgd, omnire, splatad} leverage 3D Gaussian Splatting's real-time capabilities to form Gaussian Scene Graphs, achieving fine-grained control with real-time rendering.
While these methods achieve increasingly higher quality renderings at interpolated viewpoints, they often struggle to maintain fidelity when extrapolating beyond the original training trajectories, which are crutial for downstream applications such as closed-loop simulation.

\paragraph{View Extrapolation with Generative Priors}

Recently, diffusion-based generative models have gained traction for enhancing visual quality in view extrapolation tasks\cite{3dgs-enhancer, viewextrapolator, viewcrafter}. In terms of driving scenarios, SGD \cite{sgd} pioneers by augmenting sparse view inputs using generated images.  DriveDreamer4D \cite{drivedreamer4d} employs pretrained world models with a cousin-data strategy to provide pseudo ground truths. ReconDreamer \cite{recondreamer,recondreamer++} further fine-tunes world models into online restoration networks. FreeVS \cite{freevs}, StreetCrafter and DriveX\cite{streetcrafter,drivex} use colored LiDAR projections as conditioning for diffusion-based generation. 

In contrast, our approach takes a different perspective. Rather than focusing on crafting specific generative priors or conditioning signals, we explore the design space of scene representations to enable more robust extrapolation--with or without the use of generative priors. Moreover, while prior work has largely focused on front-view or single-camera extrapolation, our method explicitly targets the more challenging multi-view extrapolation setting.

%% file: sec/3_method.tex
\section{Method}
\begin{figure*}
    \centering
    \includegraphics[width=\linewidth]{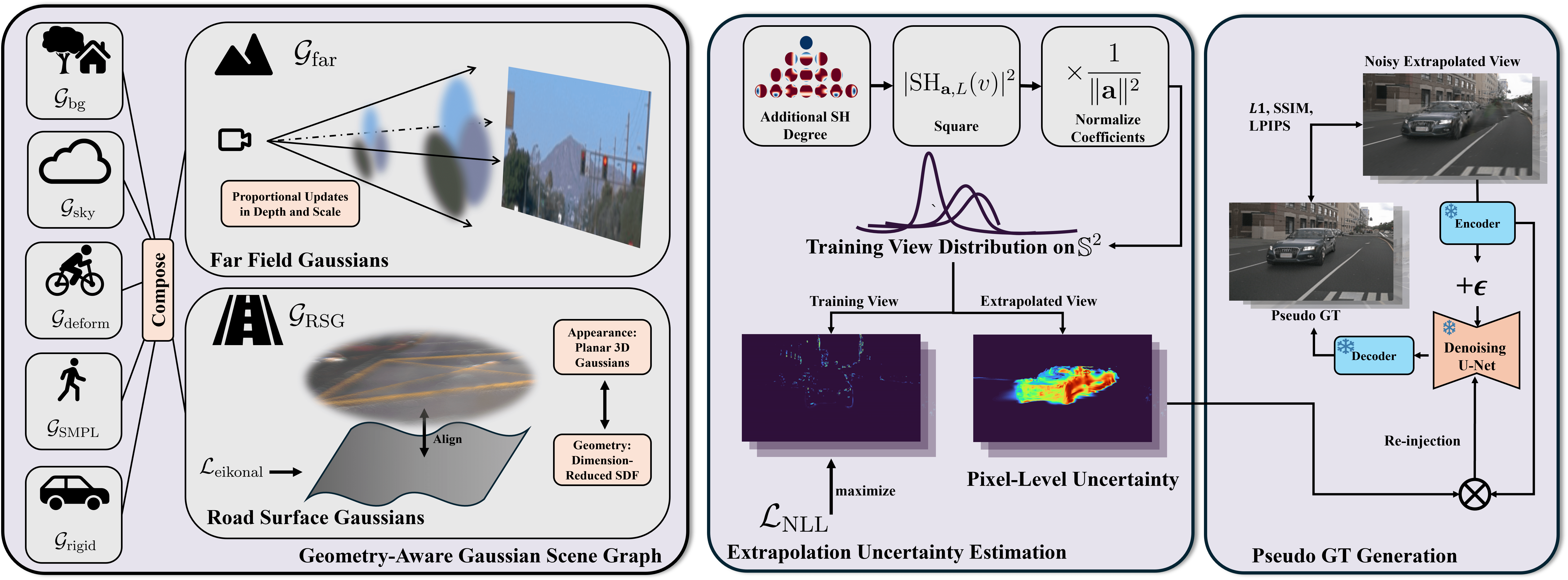}
    \caption{\textbf{Overview of ExtraGS}: \textit{Left}: ExtraGS introduces two additional node to the scene graph structure: Road Surface Gaussians (RSG) that combines 3D Gaussians with dimension-reduced SDF for road surface modeling, and Far Field Gaussians(FFG)
enabling rapid convergence for distant objects through
aligned depth-scale adjustments\textit{Middle}: ExtraGS dis-
tinguishes extrapolation-induced from scene-inherent noise,
enabling selective generative refinement of unreliable re-
gions. \textit{Right}: We leverage off-the-shelf generative models to
        generate pseudo ground truth at extrapolated views. The estimated uncertainty
        provides pixel level control to generative prior. 
        }
     \label{fig:pipeline}
\end{figure*}
\label{sec:method}
\paragraph{Problem Formulation}
Given a driving log with images $\{I_i\}_{i=1}^N$, ego poses $\{P\}_{i=1}^N$ and a set of cameras $\{K_j\}_{j=1}^C$, our objective is to learn a 4D representation that maps time, pose and camera to an image $\mathcal{S}: \mathbb{R}\times SE(3)\times C \rightarrow \mathbb{R}^{H\times W\times 3}$. Ideally, this representation enables rendering from arbitrary viewpoints and timesteps—including those beyond the original trajectory. 

\subsection{Geometric-Aware Dynamic Scene Modeling}
Following NSG~\cite{nsg}, we represent dynamic driving scenes as a scene graph where each node models elements with distinct geometric properties (e.g., infinitely far skies, rigid vehicles). To render a scene from a given viewpoint, each node is rendered independently according to its pose, and the results are composited via alpha blending.

In this section, we introduce two novel nodes, termed \textit{Road Surface Gaussians} (RSG) and \textit{Far Field Gaussians} (FFG), into the scene graph inventory. A complete description of the overall scene graph structure and other node types is provided in the supplementary material. 

\subsection{Road Surface Gaussians}
\label{sec:method_recon_road}

Reconstructing road surfaces presents unique challenges, including low-texture regions and drastic view direction changes, which often result in severe distortions under extrapolated viewpoints. To address these issues, we propose to incorporate strong geometric priors. Specifically, we assume that road surfaces are \textbf{locally planar} and exhibit \textbf{globally smooth elevation transitions}. These assumptions align well with human perceptual expectations and motivate our dimension-reduced SDF formulation. To retain fast rendering speed, we align a set of 3D Gaussians with the learned SDF to represent appearance efficiently. 


\begin{figure}
    \centering
    \includegraphics[width=0.7\linewidth]{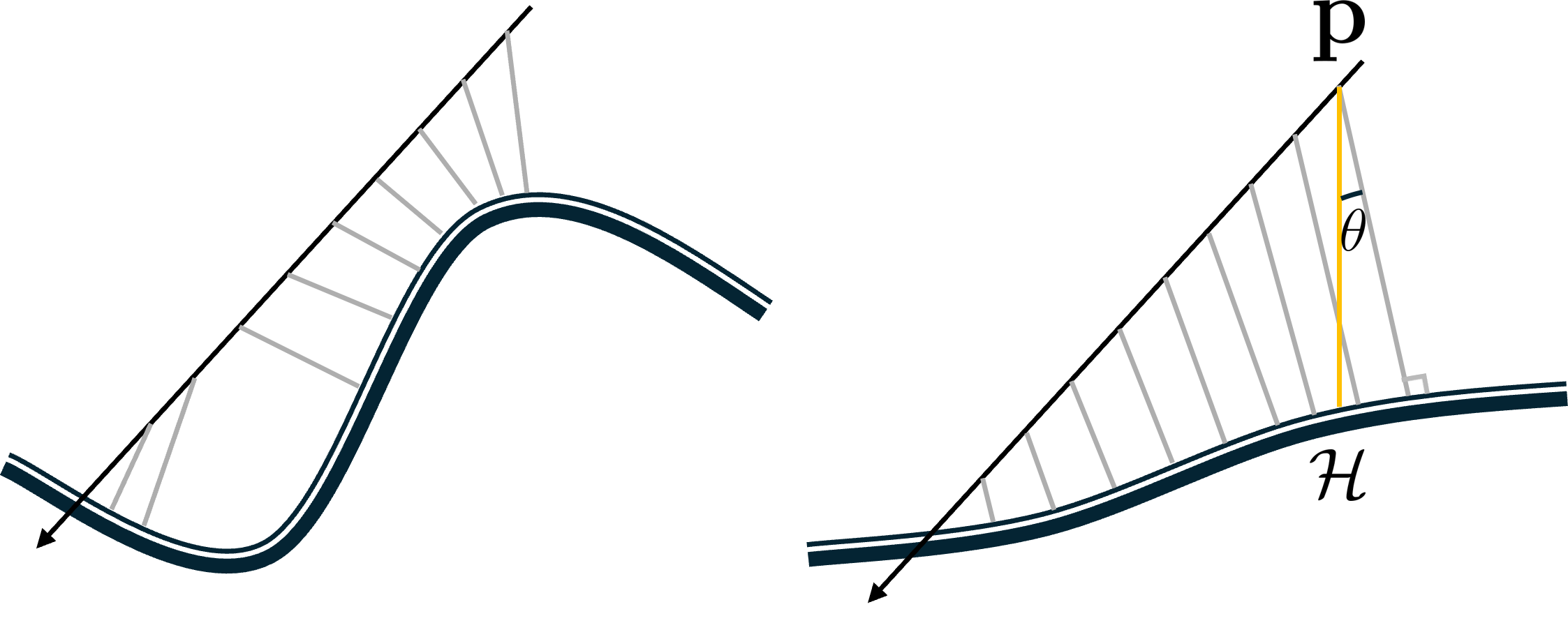}
    \caption{\textbf{Geometric motivation for dimension-reduced SDF}. Gray segments show perpendicular distances to surfaces during ray marching. \textit{Left}: General surface. \textit{Right}: Road surface with locally planar structure yielding nearly parallel distance vectors. This enables our 2D formulation $d(\mathbf{p}) = |\cos\theta|(p_{z} - \mathcal{H}(p_{x},p_{y}))$, where $|\cos\theta|$ converts vertical to perpendicular distance.}
    \label{fig:sdf}
\end{figure}
\paragraph{SDF with Planar Priors}
We propose a dimension-reduced SDF that explicitly leverages the locally planar structure of road surfaces. Our key insight is that while road geometry may curve globally, it can be well approximated as planar in the local neighborhood of a ray. This enables parameterizing the signed distance field using only 2D horizontal coordinates(Figure~\ref{fig:sdf}). 
Let $\mathbf{p} = (p_x, p_y, p_z)$ be a 3D point in a coordinate system where the $z$-axis points upward. Under the local planar assumption, the signed distance is defined as:
\begin{equation}
d(\mathbf{p}) = |\cos\theta|\left(p_{z} - \mathcal{H}(p_{x},p_{y})\right)
\end{equation}
where $\mathcal{H}: \mathbb{R}^2 \rightarrow \mathbb{R}$ represents the height of the local surface at horizontal coordinates $(p_x, p_y)$, and $\theta$ is the angle between the surface normal and the vertical axis.
Both $\mathcal{H}$ and $|\cos\theta|$ depend solely on $(p_x, p_y)$ and are parameterized using neural networks. This leads to the following 2D SDF formulation:
\begin{equation}\label{eq:2dsdf}
\left\{
\begin{aligned}
d(\mathbf{p}) &= \text{MLP}_{\text{slope}}(\mathbf{f})\left[p_z -\text{MLP}_{\text{elevation}}(\mathbf{f})\right]\\
\mathbf{c}(\mathbf{p}, \mathbf{v}) &= \text{MLP}_{\text{color}}(\mathbf{f},\mathcal{F}(\mathbf{v}))
\end{aligned}
\right.
\end{equation}
where $\mathbf{f} = \gamma(p_x, p_y)$ is a learned feature extracted from multi-level hash grids~\cite{neuralangelo}, and $\mathcal{F}$ denotes a Fourier embedding~\cite{fourier} of the view direction $\mathbf{v} \in S^2$. The functions $\text{MLP}_\text{slope}$, $\text{MLP}_\text{elevation}$, and $\text{MLP}_\text{color}$ predict the surface slope, elevation, and view-dependent color, respectively. Our 2D SDF formulation can be seamlessly integrated into the volume rendering framework from \cite{neus} to produce the rendered image $\hat{\mathbf{C}}$.

\paragraph{Road Surface Gaussians}
To take advantage of the high rendering speed of 3D Gaussians, we introduce a set of Gaussians $\mathcal{G}_{\text{RSG}}$ aligned with the learned SDF. Each Gaussian is positioned at $\boldsymbol{\mu} = (\mu_x, \mu_y, \mathcal{H}(\mu_x, \mu_y))$, where the height is queried directly from the SDF. Since road surfaces are assumed to be locally planar, each Gaussian is defined with only two scale dimensions. During training, gaussian positions are recomputed without gradient at each iteration. Only the RGB appearance attributes of these Gaussians are optimized.

\subsubsection{Optimization}
At each training iteration, we randomly sample an image $\mathbf{I} \in \mathbb{R}^{H\times W \times 3}$ from the driving log and select a batch of pixels $\mathbf{P}\in \mathbb{R}^{B\times3}$ from $\mathbf{I}$ corresponding to the road surface, using a precomputed semantic segmentation mask. The SDF is optimized using an $\mathcal{L}_1$ reconstruction loss combined with an Eikonal regularization~\cite{eikonal} to enforce smooth surface:
\begin{align}
    \mathcal{L}_\text{sdf}=\|\hat{\mathbf{P}} - \mathbf{P}\|_1 + \sum_\mathbf{\mathbf{r}\in \mathbf{P}}\sum_{i=1}^N(\|\nabla d(\mathbf{r}_i)\|_2 -1 )^2
\end{align}
Notably, the SDF acts purely as a geometric prior for Road Surface Gaussians during training. At test time, only the gaussians $\mathcal{G}_\text{RSG}$ are rendered. This design allows RSGs to benefit from the structural fidelity of the SDF representation while retaining the speed and flexibility of 3D Gaussian rendering.
\subsection{Far Field Gaussians}
When reconstructing driving scenes, distant objects span vastly different depth ranges—from buildings at hundreds of meters to clouds at tens of thousands of meters. These distant structures typically lack LiDAR coverage, making direct 3D Gaussian optimization through gradient descent on positional parameters $\boldsymbol{\mu}$ both inefficient and prone to local minima. While such suboptimal solutions may produce acceptable results along the training trajectory, they exhibit significant artifacts when rendering from extrapolated viewpoints.

To address this challenge, we introduce a new Gaussian node termed Far Field Gaussians $\mathcal{G}_\text{FFG}$ that jointly modulates both the position and scale of each Gaussian primitive (Figure~\ref{fig:pipeline}). Specifically, we augment each Gaussian with a learnable scaling factor $f$ that simultaneously transforms position $\boldsymbol{\mu}$ and scale $\boldsymbol{s}$

\begin{equation}\label{eq:dsa}
\left\{
\begin{aligned}
\boldsymbol{\mu}' &= \exp(f) \cdot \boldsymbol{\mu} \\
    \mathbf{s}' &= \exp(f) \cdot \mathbf{s}
\end{aligned}
\right.
\end{equation}
The scaling factor $f$ is jointly optimized with all other Gaussian parameters during training.

Intuitively, this formulation decouples the relative scale adjustment from absolute positioning, allowing 3D Gaussians to rapidly adapt to represent distant structures without requiring potentially ill-conditioned position updates.

\subsection{Extrapolation Uncertainty Estimation}
\label{sec:method_gen}

Recent works have demonstrated that incorporating generative priors as pseudo ground truths can significantly improve view extrapolation performance. However, a critical limitation is that generative models cannot distinguish between noise induced by viewpoint extrapolation versus scene-inherent characteristics. This leads to either meaningful components being over-smoothed (Figure~\ref{fig:demo_uncert}),  or redundant artifacts being preserved in extrapolated views (Figure~\ref{tab:ablation_uncertainty}). This motivates the need to quantify which pixels are \textit{``uncertain'': meaning a high probability of noise induced by extrapolation}. We address this through a self-supervised uncertainty estimation framework that identifies pixels likely to be corrupted by extrapolation noise.

For each pixel in an extrapolated view, we define uncertainty as the \textbf{unlikelihood} of its viewing ray $\mathbf{r}$ aligns with training viewing rays from the driving log. This can be formalized as a kernel density estimation (KDE) problem.  For each gaussian $g$, we wish to quantify the distribution $p_g(v)$ of training views, $v\in\mathbb{S}^2$. We formulate this distribution as:
\begin{align}
\label{eq:density}
    p_g(v) := \frac{1}{N}\sum_{i=1}^N K(v, v_i) \approx \frac{|\text{SH}_{\mathbf{a}, L}(v)|^2}{|\mathbf{a}|^2}
\end{align}
where $v_i$ are training viewing directions associated with $g$, and $K$ is a directional similarity measurement(e.g. cosine similarity). 

To avoid storing all training view directions explicitly, we approximate $p_g(v)$ using spherical harmonics (SH). Therefore, Equation~\ref{eq:density} is further simplified to $\frac{|\text{SH}_{\mathbf{a}, L}(v)|^2}{|\mathbf{a}|^2}$, where $\mathbf{a} \in \mathbb{R}^{(L+1)^2}$ are learnable SH coefficients and $L$ is the SH degree. This follows from SH's completeness in representing square-integrable functions on the sphere. A formal justification is provided in supplementary material.

After modeling the per-gaussian training view distribution, the uncertainty map is rendered as: $U \in \mathbb{R}^{1\times H \times W} = 1-\sum_{i=1}^N\prod_{j=1}^{i - 1} (1-\alpha_i)\alpha_i p_g$, and optimized through a log-likelihood loss:
\begin{align}
    \mathcal{L}_\text{NLL} = -\frac{1}{HW}\sum_{i,j}\log(1- U_{i,j})
\end{align}
The process is demonstrated in Figure~\ref{fig:pipeline}. Since the 3DGS pipeline already make use of spherical harmonics for view-dependent colors, Our uncertainty  can be seamlessly integrated into the Gaussian Splatting framework by storing a second set of learnable SH coefficients. 



\subsection{Pseudo Color Encoding}
In practice, we observe that pseudo ground truth images often exhibit over-saturated colors or visual artifacts. These inconsistencies result in noticeable color mismatches between the pseudo ground truth and the rendered images.
We introduce a Pseudo Color Encoding layer that learns a corrective transformation in color space:
$
C_{\text{out}} = C_{\text{in}} + \text{MLP}_{\text{pseudo}}(C_{\text{in}}),
$
where  $\text{MLP}_{\text{pseudo}}$ predicts residual color adjustments. This layer is applied only during training on extrapolated views, mitigating interference with original trajectory renderings.

\subsection{Loss Functions}
In practice, to maintain each Gassian only represent its desired category, we adopt a strategy similar to prior works~\cite{omnire, street_gs} by leveraging semantic road and sky masks predicted by a pretrained segmentation model, and a cross-entropy loss term $\mathcal{L}_{\text{mask}}$ is applied. The total loss is defined as\footnote{Relative weighting between loss terms is omitted here for brevity.}:
\begin{align*}
    \mathcal{L} = (1-\mathds{1}_\text{extrapolate}) (\mathcal{L}_1 + \mathcal{L}_\text{SSIM} + \mathcal{L}_\text{LPIPS} + \mathcal{L}_\text{mask} \\ + \mathcal{L}_\text{sdf} +
 \mathcal{L}_\text{NLL}) 
    + \mathds{1}_\text{extrapolate}\mathcal{L}_\text{ex}
\end{align*}

\section{Implementation Details}
To demonstrate the generality of ExtraGS, we integrate two off-the-shelf generative priors to our scene representation. 
\paragraph{ExtraGS-M} We first adopt MagicDriveV2~\cite{mgdv2} as the generative prior. MagicDriveV2 is a flow-matching based video generation model. We utilize its denoising process of the flow matching model to generate pseudo ground truth, where low-uncertainty pixels are gradually "re-injected" to preserve high-fidelity details. This enables selective leveraging of generative priors only where most beneficial.  The generation pseudo code in shown in the supplementary material. 

\paragraph{ExtraGS-D} We use Difix3D~\cite{difix3d+}, based on Stable Diffusion Turbo, as our generative prior. Since Difix3D uses single-step denoising, we provide pixel-level uncertainty control by downscaling Gaussian opacity according to $\alpha'=u\cdot \alpha$, effectively removing occluded objects and unwanted artifacts.

\paragraph{Experimental Setup}
We follow OmniRe\cite{omnire} for data processing and Gaussian Scene Graph implementation. In all our experiments, we first pretrain the model with 30,000 iterations, then the pseudo ground truth is introduced at iteration 30,000 and 35000 by gradually shifting laterally by 1.5m and 3m, totaling 40,000 iterations. Loss coefficients $\lambda_1, \lambda_\text{SSIM}, \lambda_\text{LPIPS}, \lambda_\text{mask}, \lambda_\text{sdf}, \lambda_\text{NLL},\lambda_\text{ex}$ are set to $0.8, 0.2, 0.05, 0.5, 0.5,1.0$ respectively. Training takes approximately 3 hours on a single NVIDIA H20 GPU. 


%% file: sec/4_experiments.tex
\section{Experiments}

\paragraph{Datasets} We evaluate our method's extrapolation performance on two large-scale autonomous driving benchmarks: nuScenes, Waymo Open Dataset(WOD)~\cite{nuscenes, waymo}. For nuScenes, all experiments are conducted at the dataset’s native resolution of 900×1600 pixels using all six surround-view cameras over 10-second clips. For WOD, we follow the protocol of~\cite{streetcrafter} by downscaling images to 1066×1600 resolution. We select scene clips approximately 4 seconds long in order to compare with previous works\cite{drivedreamer4d}. Additional experimental details are in the supplementary material.

\paragraph{Baselines} 
We compare ExtraGS-D against state-of-the-art methods that incorporate generative for extrapolation: DriveDreamer4D~\cite{drivedreamer4d}, ReconDreamer~\cite{recondreamer}, ReconDreamer++~\cite{recondreamer++}, FreeVS~\cite{freevs}, DriveX~\cite{drivex}, StreetCrafter~\cite{streetcrafter}, and Difix3D+. 
To further demonstrate ExtraGS's generality across generative prior selections and multi-camera settings, We compare ExtraGS-D  against EmerNeRF~\cite{emernerf}, PVG~\cite{pvg}, Street Gaussian~\cite{street_gs}, and OmniRe~\cite{omnire} on the nuScenes dataset.

\subsection{Downstream Tasks on the Waymo Open Dataset}

\begin{table}[htb]
\small
    \centering
    \begin{tabular}{lp{1.5cm}cc}\toprule
         Method &Gen. Condition &  NTA-IoU$\uparrow$ &NTL-IoU$\uparrow$ \\\midrule
         DriveDreamer4D &3D box + HD Map & 
       0.457& 53.30 \\
 ReconDreamer &3D box + HD Map &  0.539&54.58 \\
 ReconDreamer++ &3D box + HD Map &  0.572& 57.06 \\
 FreeVS& LiDAR Projection &  0.505&56.84 \\
 DriveX &LiDAR Projection & 0.567&\underline{58.29} \\
 StreetCrafter& LiDAR Projection& \underline{0.582}&57.15\\ 
 Difix3D+OmniRe&\textbf{None} & 0.572& 53.86\\
 ExtraGS-D&\textbf{None} & \textbf{0.592}&\textbf{58.49}\\ \bottomrule\end{tabular}
    \caption{\textbf{Downstream Tasks Performance on the Waymo Open Dataset.} ExtraGS-D achieves state-of-the-art results without requiring additional conditioning inputs.}
    \label{tab:sys_waymo}
\end{table}

\begin{figure*}
    \centering
    \includegraphics[width=1.0\linewidth]{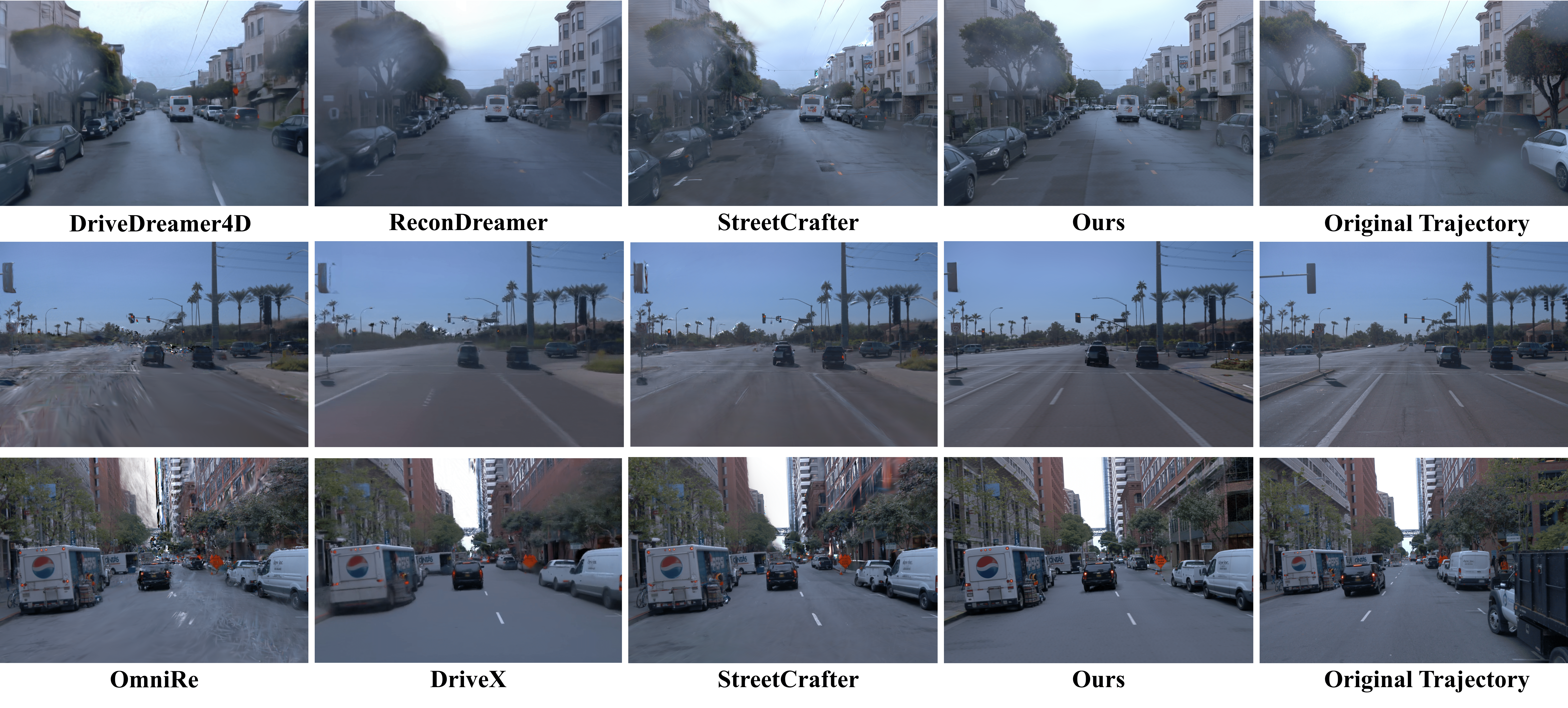}
    \caption{\textbf{Qualitative Comparisons between different method on the Waymo Open Dataset}. The results of DriveDreamer4D, ReconDreamer and DriveX are taken from their official visualizations or rendered from their provided code and weights.}
    \label{fig:waymo_qual}
\end{figure*}

\begin{table*}[t!]
\small
    
    \centering
    
    \begin{tabular}{lcccccccl}
        \toprule
         &\multicolumn{3}{c}{Extrapolation}&& \multicolumn{4}{c}{Original Trajectory}\\
        
         & FID @ 3m$\downarrow$ & $\text{mAP}_{obj}$$\uparrow$& $\text{mAP}_{map}$ $\uparrow$ 
& &PSNR$\uparrow$ & SSIM$\uparrow$& LPIPS$\downarrow$ 
 & FID$\downarrow$ \\
        \midrule
        EmerNerf& 104.49& 0.137& 0.217& & 28.82& 0.826& 0.390 & 51.75\\
        PVG& 138.34& 0.319& 0.164& & 28.95& 0.846&  0.332& 60.04\\
        StreetGS& 95.29& 0.450& 0.321& & 29.64& 0.878&  0.240& 26.85\\
        OmniRe& 93.60& 0.419& 0.308& & 29.70& \textbf{0.879}&  \textbf{0.236}& 26.74\\
 ExtraGS-M& \textbf{77.19}& \textbf{0.513}& \textbf{0.384}& & \textbf{29.72}& 0.874& 0.242&\textbf{26.30}\\
 \bottomrule
        \end{tabular}
    
    \caption{\textbf{System Level Comparison on the nuScenes dataset.} Results are reported with resolution 900x1600 across all 6 cameras. Our method achieves enhanced extrapolation results while maintaining visual quality in the original trajectory. 
    }
    \label{tab:nuscenes}
\end{table*}
We evaluate ExtraGS-D's practical utility through downstream perception tasks on the Waymo Open Dataset, comparing against state-of-the-art trajectory extrapolation methods. The results are shown in Table~\ref{tab:sys_waymo} and Figure~\ref{fig:waymo_qual}. 
\paragraph{Evaluation Protocol} We employ the NTA-IoU and NTL-IoU scores defined in DriveDreamer4D with their offical implementation. NTA-IoU measures object detection performance by computing the IoU between YOLO11~\cite{yolo11} predictions and ground truth 3D bounding boxe projections. NTL-IoU assesses lane detection performance through IoU between TwinLiteNet~\cite{twin} predictions and ground truth HD maps. All extrapolation trajectories are laterally shifted by 3 meters to simulate lane-shift scenarios.

\paragraph{Results}We compare against seven state-of-the-art methods with varying conditioning requirements. Results for DriveDreamer4D, ReconDreamer, FreeVS, and DriveX are directly taken from \cite{recondreamer, recondreamer++, drivex}. We use official implementations for StreetCrafter and integrate Difix3D with OmniRe for dynamic object handling.

ExtraGS-D achieves state-of-the-art performance with NTA-IoU of 0.592 and NTL-IoU of 58.49, surpassing all competing methods. Notably, ExtraGS-D does not require additional conditioning inputs. The comparison with Difix3D+OmniRe (NTA-IoU: 0.572, NTL-IoU: 53.86) further validates ExtraGS's effectiveness, as using the same generative prior (Difix3D) with a different scene representation produces significantly inferior results.

Figure~\ref{fig:waymo_qual} demonstrates ExtraGS-D's superior visual quality across diverse scenarios. ExtraGS-D consistently generates crisp lane markings while maintaining visual coherence with the original trajectory, whereas competing methods often produce blurry or oversaturated lane markings. 

\subsection{System-Level Comparison on the nuScenes Dataset}


To further demonstrate the generality of ExtraGS, we present results using a pre-trained video generation model MagicDriveDiT as the generative prior(ExtraGS-M). The results are show in Table~\ref{tab:nuscenes}.

\paragraph{Evaluation Protocol} We assess performance across two categories: extrapolation quality using Fréchet Inception Distance(FID) and downstream task accuracy via multi-camera 3D object detaction($\text{mAP}_{obj}$) and HD map prediction ($\text{mAP}_{map}$), and original trajectory fidelity using standard metrics(PSNR, SSIM and LPIPS)report results of the commonly adopted PSNR, SSIM and LPIPS on the original trajectory. For downstream evaluations, we employ pretrained BEVFormer V2~\cite{bevformerv2} and MapTR V2~\cite{maptrv2} to predict at the extrapolated views. The predictions are shifted back to the original trajectory for evaluation. We report average precision (AP) for 3D object detection (focusing on the “car” category) and mean average precision (mAP) for map prediction across chamfer distance thresholds of 0.5, 1.0, and 1.5 meters. 

\paragraph{Analysis}  Table~\ref{tab:nuscenes} shows ExtraGS-M achieves superior extrapolation performance with the lowest FID@3m (77.19) and highest downstream task scores ($\text{mAP}_{obj}:0.513,\text{mAP}_{map}:0.384$), significantly outperforming all baselines. Critically, these improvements come without degrading original trajectory quality—ExtraGS-M maintains competitive PSNR (29.72) and FID (26.30) scores. 

\subsection{Ablation Studies}

\label{sec:exp_ablation}
\begin{figure*}[ht]
    \centering
    \includegraphics[width=1.0\linewidth]{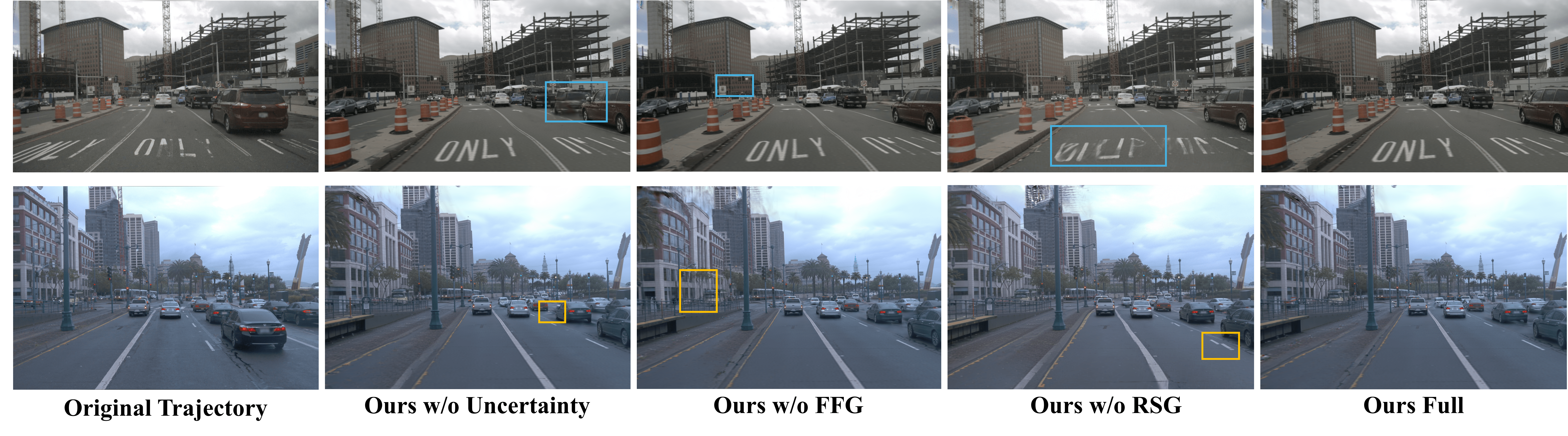}
    \caption{\textbf{Visual Ablation on Key Components of ExtraGS}}
    \label{fig:ablation_qual}
\end{figure*}

\paragraph{Road Surface Representation}
\begin{table}[]
\small
\centering

\begin{tabular}{lccc} 
\toprule
Method & PSNR$\uparrow$&SSIM$\uparrow$& FID$\downarrow$ @ 3m \\
\midrule
3DGS\cite{3dgs}& \textbf{29.22} &\textbf{0.862}& 90.67\\
2DGS\cite{2dgs}& 24.67 &0.758& \underline{75.51}\\
SDF\cite{streetsurf}& 27.81 &0.741& 77.08\\
Ours & \underline{27.92} &\underline{0.812}& \textbf{67.72}\\
\bottomrule
\end{tabular}
\caption{\textbf{Ablation study on road surface representation}}
\label{tab:ablation_road_surface}
\end{table}

\begin{figure}
    \centering
    \includegraphics[width=\linewidth]{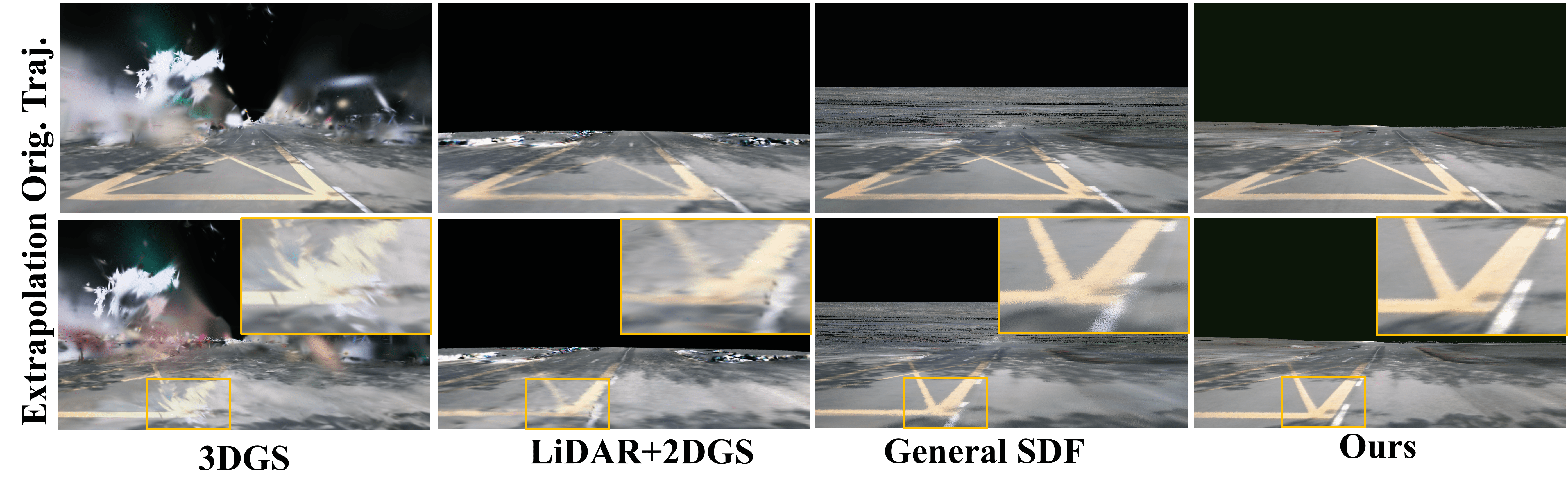}
    \caption{\textbf{Qualitative Ablations on Different Road Surface Representations}}
    \label{fig:qual_road}
\end{figure}

We investigate different road surface modeling strategies \textit{without} generative priors(Table~\ref{tab:ablation_road_surface} and Figure~\ref{fig:qual_road}). We compare 3DGS, 2DGS with LiDAR initialization~\cite{2dgs}, SDF based on StreetSurf~\cite{streetsurf}, and our RSG representation. We evaluate reconstruction quality using PSNR and SSIM computed on drivable areas indicated by preprocessed semantic road masks, and assess perceptual consistency under extrapolation using FID at 3-meter lateral shifts.

The results reveal a fundamental trade-off between reconstruction fidelity and extrapolation quality. While 3DGS achieves the highest PSNR (29.22) and SSIM (0.862), it fails to capture accurate road surface geometry, resulting in poor extrapolated view quality (FID: 90.67). This limitation stems from 3DGS's inability to enforce geometric constraints on road surfaces. In contrast, our proposed RSG model achieves the optimal balance, delivering the lowest extrapolation FID (67.72) while maintaining competitive reconstruction quality, demonstrating effective geometric modeling for road surfaces.


\paragraph{Component Analysis}
To understand how each component contributes to the final extrapolation quality, we present a comprehensive ablation in Table~\ref{tab:ablation_uncertainty} and Figure~\ref{fig:ablation_qual}.

\begin{enumerate}
    \item[-] \textbf{Road Surface Gaussians (RSG)}: Removing RSG results in notable performance degradation even \textit{with} generative priors (FID increases from 81.03 to 84.74). Visual results in Figure~\ref{fig:ablation_qual} (column 4) reveal significant distortions in lane markings, demonstrating the complementary roles of geometric and generative priors in trajectory extrapolation. 
    \item[-] \textbf{Uncertainty Estimation}: The absence of uncertainty guidance leads to degraded performance across both original trajectory quality and extrapolation consistency (FID: 87.47). Figure~\ref{fig:ablation_qual} (column 2) shows that without uncertainty, artifacts appear in areas occluded in the original trajectory. 
    \item[-] \textbf{Far Field Gaussians(FFG)}: Removing FFG Node significantly impacts both original trajectory PSNR and extrapolation FID. Further, this degradation cannot be resolved using the generative prior, as shown in Table~\ref{tab:ablation_uncertainty} (column 3) and Figure~\ref{fig:ablation_qual}. 
    \item[-] \textbf{Pseudo Color Encoding}: The removal of pseudo color encoding causes a substantial drop in original trajectory PSNR (from 32.31 to 30.86). We believe this degradation occurs due to the color biases in the generated images. 
\end{enumerate}

\begin{table}[]
\centering
\small
\setlength{\tabcolsep}{2.1pt}
\begin{tabular}{lccc} 
\toprule
& PSNR$\uparrow$&FID (before)$\downarrow$& FID (after)$\downarrow$\\
\cmidrule{2-4}
Ours w/o FFG& 31.15&147.85& 87.19\\
 Ours w/o RSG& 31.95& 138.92&84.74\\
Ours w/o Uncertainty& 31.94&--& 87.47\\
Ours w/o Pseudo Color Encod.& 30.86&--& 85.20\\
Ours Full& \textbf{32.31}&\textbf{128.36}& \textbf{81.03}\\
\bottomrule
\end{tabular}
\caption{\textbf{Ablation of Key Components} We report PSNR in the original trajectory, and extrapolation FID before and after applying pseudo ground truths.}
\label{tab:ablation_uncertainty}
\end{table}

%% file: sec/5_conclusion.tex
\section{Conclusion}

We present ExtraGS, a holistic framework integrating geometric and generative priors for extrapolated view synthesis. Built within the NSG paradigm, ExtraGS introduces two Gaussian nodes: Road Surface Gaussians (RSG) combining 3D Gaussians with dimension-reduced SDF for road surface modeling, and Far Field Gaussians(FFG) enabling rapid convergence for distant objects through aligned depth-scale adjustments. Additionally, ExtraGS distinguishes extrapolation-induced from scene-inherent noise, enabling selective generative refinement of unreliable regions.
Extensive experiments demonstrate superior perceptual quality and structural consistency in extrapolated views. 





